\documentclass[preprint,review,times,10pt]{elsarticle}
\usepackage{amsmath}
\usepackage{amssymb}
\usepackage[hidelinks]{hyperref}
\usepackage{cleveref}
\usepackage{booktabs}
\usepackage{multirow}
\usepackage{graphicx}
\usepackage{adjustbox}
\usepackage{float}
\usepackage{pifont}    
\usepackage{threeparttable}
\usepackage{tabularx}   
\usepackage{enumitem}   
\usepackage{ragged2e}   
\usepackage{xurl}
\usepackage{xcolor}
\usepackage{wrapfig}

\journal{Pattern Recognition}

\begin{document}

\begin{frontmatter}

\title{DisasterBench: A Multimodal Benchmark for UAV-Based Disaster Response in Complex Environments}

\author[a]{Tan Zhang} 
\author[a]{Quanyou Li}
\author[b]{Lu Zhang}
\author[c]{Jun Liu}
\author[d]{Xiaofeng Zhu}
\author[a]{Ping Hu\corref{cor1}}

\cortext[cor1]{Corresponding author}
\ead{pinghu@uestc.edu.cn}

\affiliation[a]{organization={School of Computer Science and Engineering, University of Electronic Science and Technology of China},
            city={Chengdu},
            postcode={611731}, 
            country={China}}
\affiliation[b]{organization={School of Information and Communication Engineering, Dalian University of Technology},
            city={Dalian},
            postcode={116024}, 
            country={China}}
\affiliation[c]{organization={School of Computing and Communications, Lancaster University},
            city={Lancaster},
            postcode={LA1 4YW}, 
            country={England}}
\affiliation[d]{organization={School of Computer Science and Technology, Hainan University},
            city={Haikou},
            postcode={570228}, 
            country={China}}

\begin{abstract}
When a disaster unfolds, responders must answer not only what is happening, but also why it is happening, what will happen next, and what to do now, often from noisy low-altitude UAV views and under tight on-site compute constraints. However, most existing multimodal benchmarks emphasize perception (e.g., recognition/description), cover limited disaster types, and provide insufficient support for the multi-stage reasoning required in practical emergency response. We introduce DisasterBench, a multi-stage multimodal reasoning benchmark for UAV-Based disaster response in complex environments. DisasterBench spans 14 disaster-related scene types and 9 response-critical tasks across pre-, during-, and post-disaster stages, with fine-grained disaster-task mappings that explicitly test causal attribution, propagation prediction, damage analysis, and decision-oriented reasoning. To enable reasoning on the edge, we further propose DisasterVL, a lightweight multimodal model optimized with a three-stage pipeline combining domain instruction tuning, chain-of-thought-guided multimodal alignment, and reinforcement learning-based policy optimization. Experiments across 21 popular MLLMs show that our 2B-parameter DisasterVL outperforms all evaluated open-source models and substantially narrows the gap to state-of-the-art closed-source models, achieving GPT-4o-comparable reasoning accuracy with superior efficiency. The project page is available at \url{https://github.com/TanmouTT/DisasterBench}.
\end{abstract}


\end{frontmatter}

\section{Introduction}
\label{sec:intro}

Disasters pose severe threats to human life and critical infrastructure, demanding rapid and reliable emergency response under highly uncertain conditions. In real-world scenarios, responders must go beyond recognizing visible damage and answer a sequence of interdependent questions: \emph{what} is happening, \emph{why} it is happening, \emph{what} will happen next, and \emph{what} actions should be taken now. These decisions are increasingly informed by low-altitude unmanned aerial vehicle (UAV) imagery, which provides timely and close-range observations in complex terrains and hazardous environments~\cite{SHIANIOS2025104318}. However, such observations are inherently noisy, partial, and affected by severe viewpoint changes and occlusions~\cite{TOWNSELL2025104561}, making effective disaster response fundamentally a problem of structured, multi-step reasoning over multimodal evidence rather than isolated perception.

Multimodal large language models (MLLMs) have shown promise in disaster analysis through cross-modal fusion and semantic modeling~\cite{yu-etal-2025-wximpactbench,lei-etal-2025-harnessing,XU2026104753}. Meanwhile, low-altitude UAVs have emerged as practical edge-sensing platforms capable of deploying lightweight models~\cite{cleland2025cognitive,li2025segearth}. Despite these advances, existing disaster-related benchmarks remain largely perception-oriented, focusing on recognition or description tasks and covering only limited disaster types. Moreover, most benchmarks treat analytical tasks in isolation and ignore how reasoning requirements evolve across different disaster stages, limiting their ability to reflect emergency response in complex environments.

We argue that disaster response in complex environments should be formulated as a multimodal reasoning problem. In practice, disaster analysis unfolds across interconnected stages, including pre-disaster risk assessment, during-disaster situational understanding, and post-disaster evaluation and decision-making, each imposing distinct reasoning constraints. The type and phase of a disaster jointly determine the available visual evidence and task priorities, requiring explicit alignment between disaster conditions and analytical objectives~\cite{yin-etal-2025-disastir}. Without modeling this coupling, evaluations risk overlooking the core reasoning challenges encountered in real emergency scenarios.

\begin{wrapfigure}{r}{0.5\textwidth}
\centering
\includegraphics[width=\linewidth]{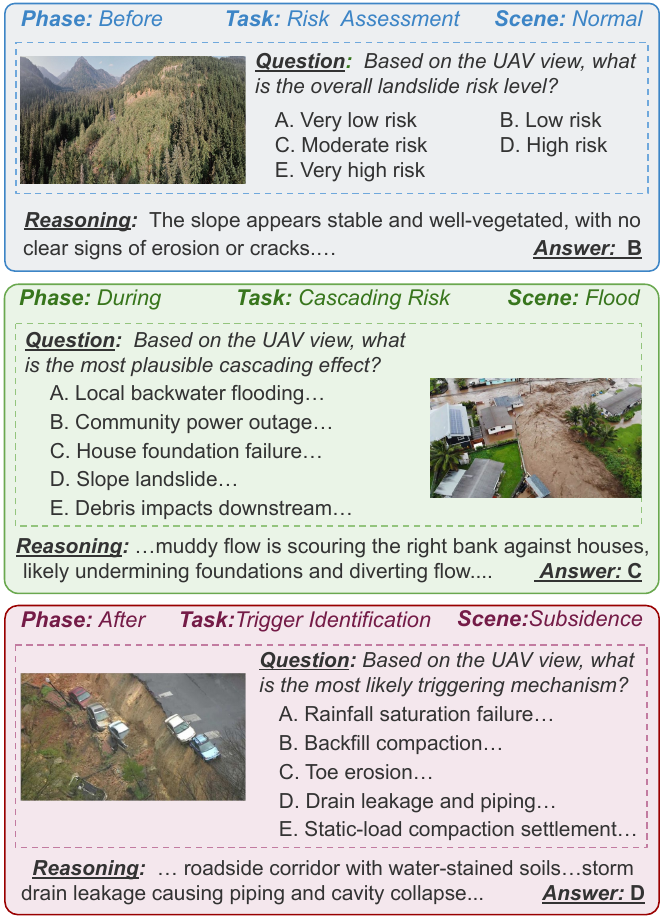}
\caption{Examples from \textbf{DisasterBench} for reasoning-oriented VQA from low-altitude UAV views across pre-, during-, and post-disaster phases. Only questions, options, brief cues, and final answers are shown.}
\label{fig:teaser}
\end{wrapfigure}

Low-altitude UAV views further amplify these challenges by introducing a fundamentally different observation setting from satellite~\cite{SHIANIOS2025104318,10.1145/3746027.3754559,10.1145/3746027.3754950} or high-altitude imagery. While overhead views provide macro-level spatial patterns, UAV imagery captures localized and fine-grained details that are frequently affected by occlusion and partial observability~\cite{10.1145/3746027.3754994}. Critical cues such as unstable terrain, blocked infrastructure, or early signs of cascading hazards may not be directly visible and must be inferred through contextual reasoning grounded in domain knowledge and physical constraints~\cite{kulahara2025can}. These characteristics make low-altitude UAV imagery a realistic yet challenging modality for evaluating multimodal disaster reasoning.

Beyond perceptual uncertainty, real disaster scenarios impose strict computational constraints. Emergency operations are often conducted with limited connectivity, power supply, and real-time requirements, making reliance on large cloud-based models impractical. Consequently, disaster reasoning systems must operate with lightweight multimodal models deployed at the edge. However, many existing approaches depend on zero-shot or prompt-based reasoning~\cite{yu-etal-2025-wximpactbench,diallo2025response,emami2025prompts} or resource-intensive pipelines~\cite{wang2025disasterm,kuai2025knowledge,shabbir2025thinkgeo}, which are incompatible with such constraints. Reliable disaster reasoning therefore requires training paradigms that enhance reasoning robustness and generalization within small model budgets.

To address these challenges, we introduce \textbf{DisasterBench}, a multi-stage multimodal reasoning benchmark for UAV-Based disaster response in complex environments. DisasterBench is constructed from \textbf{5,330} real-world low-altitude UAV images and comprises \textbf{29,300} reasoning-oriented samples, covering 14 disaster-related scene types and 9 response-critical tasks spanning pre-disaster, during-disaster, and post-disaster stages. To make such reasoning capabilities practical, we further propose \textbf{DisasterVL} trained via a lightweight-model optimization pipeline that integrates domain-specific instruction tuning, chain-of-thought-guided multimodal alignment, and reinforcement learning-based policy optimization. Together, DisasterBench and the proposed framework enable systematic evaluation and effective learning of multimodal disaster reasoning under realistic observational and computational constraints. In summary, our main contributions are as follows:
\begin{itemize}
    \item We introduce \textbf{DisasterBench}, a multi-stage multimodal reasoning benchmark for UAV-Based disaster response in complex environments, covering diverse disaster types, response-critical tasks, and all pre-/during-/post-disaster stages.
    \item We propose \textbf{DisasterVL} with an optimization recipe for lightweight multimodal models, combining domain knowledge injection, CoT-guided multimodal alignment, and reinforcement learning, to improve reasoning robustness under limited compute.
    \item We benchmark \textbf{21} popular multimodal models (closed- and open-source) and show that our trained DisasterVL with 2B parameters achieves strong, balanced performance on DisasterBench, substantially narrowing the gap to state-of-the-art closed-source models.
\end{itemize}

\begin{table*}[t]
    \centering
    % \vspace{-0.3cm}
    \label{tab:comparison}
    \resizebox{\textwidth}{!}{
    \begin{tabular}{lcccccc}
        \toprule
        \textbf{Dataset} &
        \textbf{View} &
        \textbf{\#Disaster} &
        \textbf{Stage} &
        \textbf{Size (Imgs/Samples)} &
        \textbf{Reasoning Target} \\
        \midrule
        RSCC~\cite{chen2025rscc}                   & Satellite & 6  & Post & 124,702 / 62,351  & Change-oriented description \\
        MONITRS~\cite{revankar2025monitrs}         & Satellite & 10 & Post     & --~~~ / 54,504  &  Change-oriented monitoring \\
        DisasterM3~\cite{wang2025disasterm}       & Satellite & 10 & Post & 26,988 / 123,010 & Change-oriented perception \\
        \midrule
        DisasterEye~\cite{jankovic2025uav}         & UAV       & 8  & During   & 2,751 / 2,751    & Disaster Classification \\
        AIDER~\cite{kyrkou2020emergencynet}         & UAV       & 5  & During   &  6,923 /  6,923    & Disaster Classification \\
        FloodNet~\cite{rahnemoonfar2021floodnet}   & UAV       & 1  & Post     & 3,200 / 11,000    & Damage perception\\
        RescueNet~\cite{rahnemoonfar2023rescuenet} & UAV       & 1  & Post     & 4,494 / 4,494    & Damage segmentation \\
        \midrule
        \textbf{DisasterBench (Ours)} &
        \textbf{UAV} &
        \textbf{14} &
        \textbf{Multi-stage} &
        \textbf{5,330 / 29,300} &
        \textbf{Causal and decision-oriented reasoning} \\
        \bottomrule
    \end{tabular}
    }
    \caption{Comparison of \textbf{DisasterBench} with representative multimodal aerial-view disaster-related datasets.}
    \label{tab:bench_previous}
\end{table*}

\section{Related Works}
Early multimodal benchmarks primarily focused on single tasks such as image captioning~\cite{lin2014microsoft,plummer2015flickr30k}, visual grounding~\cite{kazemzadeh2014referitgame}, visual question answering~\cite{marino2019ok,mathew2021docvqa,schwenk2022okvqa}, and optical character recognition~\cite{singh2021textocr}, mainly evaluating perception and basic semantic understanding. With the rapid development of large-scale vision--language models, recent benchmarks have evolved toward more complex and diverse evaluations. These benchmarks broadly fall into two categories: domain-specific benchmarks targeting specialized reasoning abilities (e.g., scientific or mathematical reasoning~\cite{yue2024mmmu,lu2024mathvista}, chart understanding~\cite{hiippala2021ai2d,masry-etal-2022-chartqa}, or hallucination detection~\cite{liu2023visual}), and general-purpose benchmarks emphasizing broad task and modality coverage, such as MME~\cite{fu2025mme}, MMBench~\cite{liu2024mmbench}, MMVet~\cite{yu2023mm}, MMStar~\cite{chen2024are}, LLaVABench~\cite{liu2023visual}, VisIT-Bench~\cite{bitton2023visit}, and TouchStone~\cite{bai2023touchstone}. 
Overall, these benchmarks broaden evaluation from early perception-centric tasks to more diverse multimodal understanding and reasoning, and are widely used to compare general-purpose MLLMs~\cite{liu2023visual,li2024survey2,huang2024survey}.

Recent years have seen increasing interest in multimodal datasets for disaster analysis and response. DisasterEye~\cite{jankovic2025uav} integrates UAV imagery with first-person views to improve robustness across disaster and non-disaster scenes. FloodNet~\cite{rahnemoonfar2021floodnet} provides post-flood imagery with classification, segmentation, and VQA tasks, while DisasterQA~\cite{rawat2024disasterqa} evaluates disaster-related knowledge through multiple-choice questions. Beyond vision-centric datasets, DisastIR~\cite{yin-etal-2025-disastir} and WXIMPACTBENCH~\cite{yu-etal-2025-wximpactbench} focus on text-based impact analysis and retrieval-oriented tasks, respectively, highlighting complementary directions for disaster understanding. On the visual side, FireSentry~\cite{zhou2025firesentry} targets wildfire monitoring with multi-sensor videos and telemetry for spatiotemporal prediction, and AIFloodSense~\cite{simantiris2025aifloodsense} addresses flood scene understanding from aerial imagery. Remote-sensing benchmarks such as RSCC~\cite{chen2025rscc} and MONITRS~\cite{revankar2025monitrs} incorporate temporal satellite imagery with language annotations to study change description and disaster evolution, and DisasterM3~\cite{wang2025disasterm} further explores multi-task instruction tuning with multimodal remote-sensing data. Table~\ref{tab:bench_previous} summarizes a systematic comparison between existing disaster-related benchmarks and our proposed dataset.
While these efforts advance multimodal perception, spatiotemporal modeling, and task diversity, most datasets emphasize specific disaster stages (often pre/post), limited disaster coverage, or perception-oriented objectives, and rarely enforce explicit coupling between disaster conditions and task requirements. In contrast, our benchmark is designed to evaluate \emph{multi-stage}, stage-aware task reasoning from low-altitude UAV views, explicitly modeling disaster evolution, task dependencies, and decision-oriented reasoning under realistic emergency-response constraints.

\section{DisasterBench}
\label{sec:benchmark}

\begin{figure*}[t]
    \centering
    \includegraphics[width=1\linewidth]{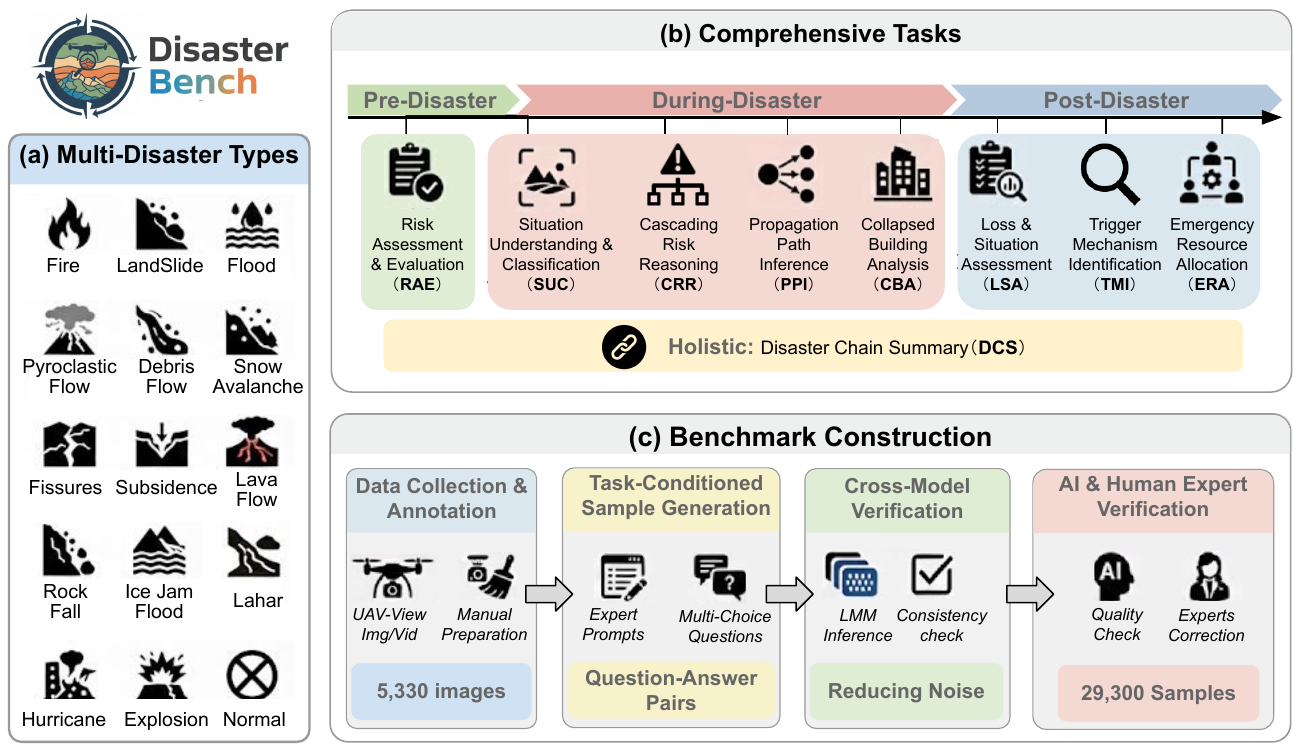}
    \caption{Overview of \textbf{DisasterBench}. (a) Disaster-related scene categories from low-altitude UAV views.  (b) comprehensive task taxonomy spanning pre-, during-, and post-disaster stages, together with a holistic disaster-chain summary task. (c) Benchmark construction pipeline, including data collection, task-conditioned sample generation, cross-model verification, and expert review.}
    \label{fig:overview}
\end{figure*}

In this section, we introduce \textbf{DisasterBench}, a multimodal reasoning benchmark for UAV-Based disaster response in complex environments, the multi-disaster types, comprehensive tasks and benchmark construction are shown in \Cref{fig:overview}. Unlike existing benchmarks that mainly evaluate perception-level capabilities, DisasterBench targets higher-level reasoning required in real emergency scenarios, including causal analysis, hazard evolution understanding, and decision-oriented assessment across the entire disaster lifecycle. The benchmark spans \textbf{14 disaster-related scene types} and defines \textbf{9 response-critical tasks} covering the \emph{pre-disaster}, \emph{during-disaster}, and \emph{post-disaster} stages, together with a holistic task for multi-stage disaster reasoning.

DisasterBench is built from \textbf{5,330} real-world low-altitude UAV images and comprises \textbf{29,300} reasoning-oriented samples in a unified multiple-choice VQA format. Each image is annotated with a single disaster label, while supporting multiple task queries conditioned on the disaster type and its current stage, enabling structured and stage-aware evaluation of multimodal disaster reasoning.

\subsection{Benchmark Composition}
\begin{figure}
    \centering
    \includegraphics[width=1\linewidth]{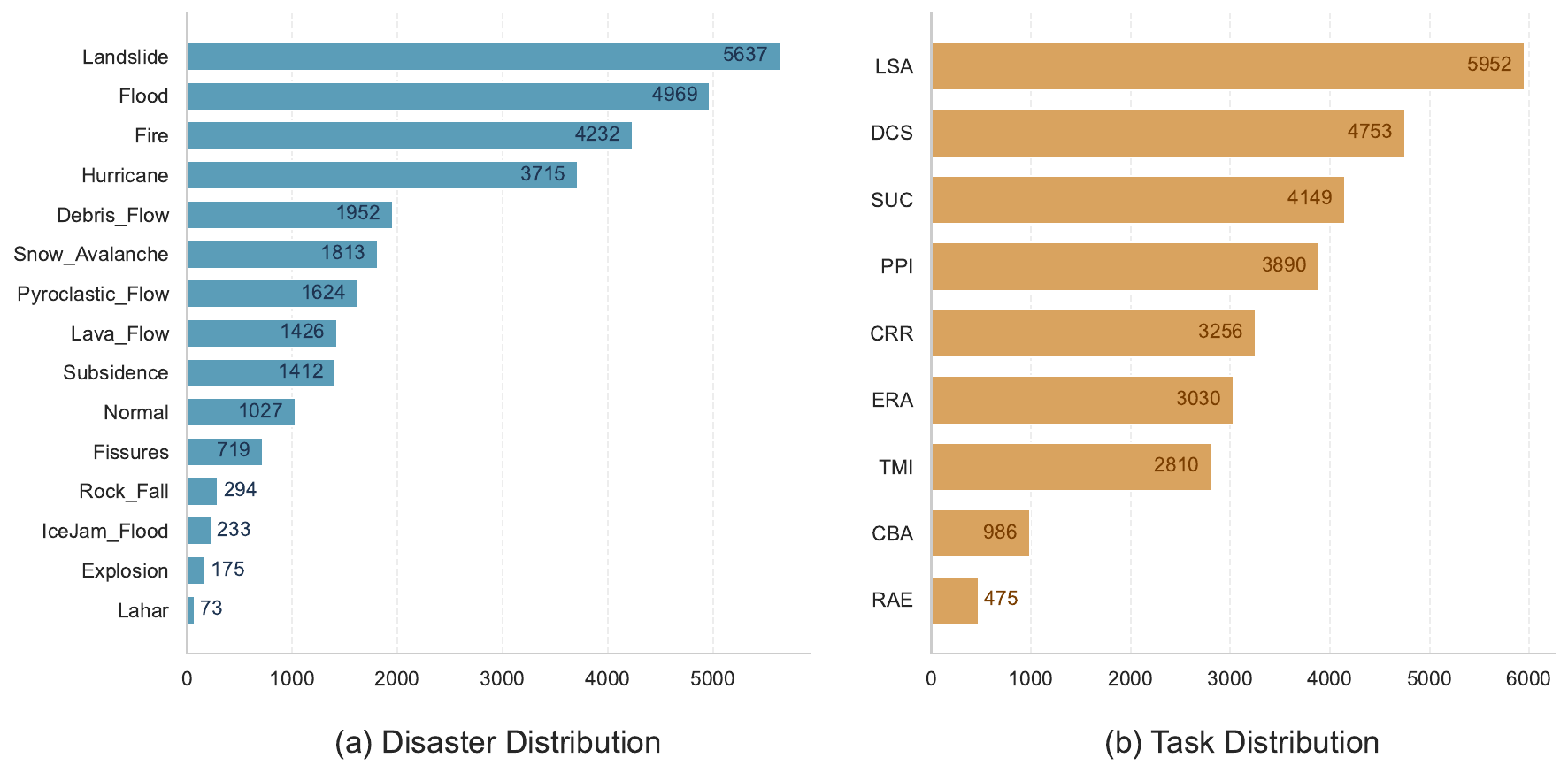}
    \caption{Sample distribution in DisasterBench.}
    \label{fig:statistics}
\end{figure}

\paragraph{Disaster categories}
As shown in Fig.~\ref{fig:overview} (a), DisasterBench includes 14 disaster-related UAV scene categories that capture hazardous processes, damage states, and geophysical phenomena relevant to emergency response: \emph{Fire, Landslide, Flood, Pyroclastic Flow, Debris Flow, Snow Avalanche, Fissures, Subsidence, Lava Flow, Rock Fall, Ice Jam Flood, Lahar, Hurricane,} and \emph{Explosion}. 
Rather than restricting categories to strictly defined disaster taxonomies, this design reflects the diverse scene conditions encountered by low-altitude UAVs during disaster operations~\cite{khan2022emerging,UNDRR_ISC_2020_HazardReview}.
The categories exhibit varied visual signatures and underlying physical or structural mechanisms, enabling evaluation of both category-specific reasoning and cross-category generalization. To support pre-disaster reasoning and robustness, we additionally include a \textit{Normal} category of non-disaster scenes.
The number of low-altitude UAV images corresponding to each disaster type is shown in Fig.~\ref{fig:statistics} (a).  

\paragraph{Comprehensive tasks}
As shown in Fig.~\ref{fig:overview} (b), we define nine tasks to capture the key reasoning demands of real-world disaster response across the full lifecycle~\cite{UNDRR2017DisasterRiskManagement}. 
In the \emph{pre-disaster} stage, models perform \emph{risk assessment and evaluation} (RAE). In the \emph{during-disaster} stage, we evaluate \emph{situation understanding and classification} (SUC),  \emph{cascading risk reasoning} (CRR),  \emph{propagation path inference} (PPI), and \emph{collapsed building analysis} (CBA). In the \emph{post-disaster} stage, tasks cover \emph{loss and situation assessment} (LSA), \emph{trigger mechanism identification} (TMI), and \emph{emergency resource allocation} (ERA). Finally, a holistic task, \emph{disaster-chain summary} (DCS), asks models to integrate observations into a structured narrative with explicit causal relations. Together, these tasks evaluate stage-aware disaster reasoning beyond isolated perceptual skills. The sample distribution for different tasks is shown in Fig.~\ref{fig:statistics} (b).

\subsection{Benchmark Construction}
DisasterBench is built via a four-stage pipeline (Fig.~\ref{fig:overview} (c)) to ensure realism, consistency, and high annotation fidelity. 
First, we collect low-altitude UAV imagery from diverse real-world sources including public disaster reports, news media footage and open video-sharing platforms. Then, we perform cleaning, deduplication, and manual category annotation to obtain a reliable set of disaster-related scenes. 

Second, we generate task-conditioned multiple-choice questions using carefully designed prompts aligned with each reasoning task. 
Since a single scene may support multiple response objectives, we construct structured disaster-task pairs and generate corresponding queries accordingly. 
Each generated question undergoes cross-model verification using multiple strong vision--language models (e.g., GPT-5, Gemini-2.5-Pro), and only samples with consistent answers across models are retained, which reduces annotation noise while preserving challenging examples for rare or complex disaster scenarios.

\begin{figure}[t]
    \centering
    \includegraphics[width=1\linewidth]{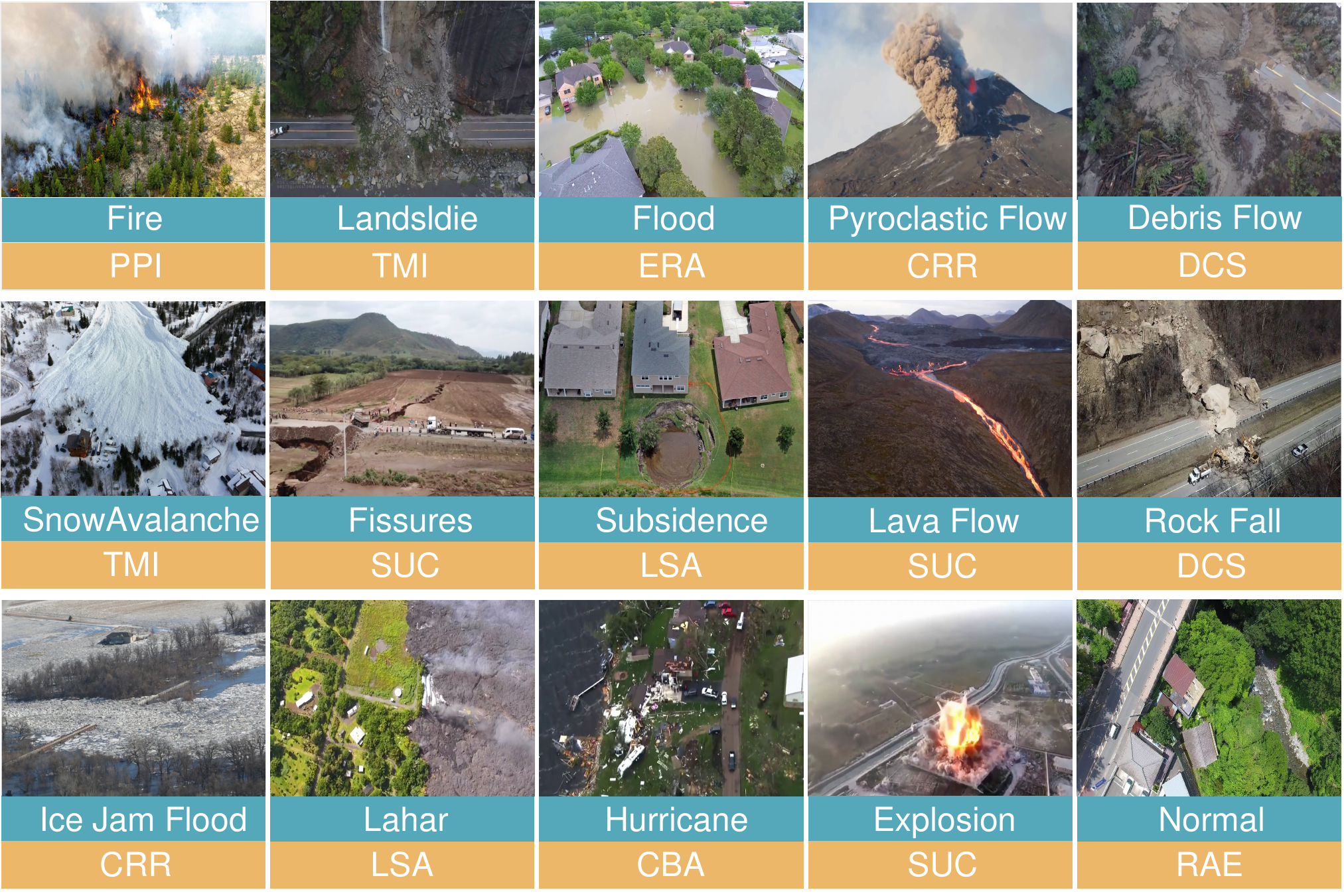}
    \caption{Representative images from DisasterBench illustrating diverse disaster scenes and related tasks.}
    \label{fig:vis_example}
\end{figure}

Third, we apply a cross-model verification procedure: multiple strong vision--language models independently answer each question, and we retain only samples whose answers are consistent across models, reducing ambiguity and annotation noise. 
Flagged or conflicting items are further audited by the experts, and any unresolved cases are discarded, ensuring high inter-annotator agreement and dataset integrity.

Finally, we conduct AI-assisted quality checks followed by expert human verification by two full-time annotators with a disaster response background. Over approximately three months, these experts review all collected images, flagging ambiguous or low-quality samples for revision or removal, which results in about 17\% of generated items being revised or discarded. This process additionally monitors representation across all disaster categories, including rare and tail-class events, to ensure that the benchmark provides comprehensive coverage and supports robust generalization. Through this, we curate approximately 5,330 real-world low-altitude UAV images and construct 29,300 high-quality reasoning samples. To provide further insight into the dataset, Fig.~\ref{fig:vis_example} presents representative category--task pairs.

\subsection{Evaluation Protocol}
All samples in DisasterBench follow a unified multiple-choice VQA format to support scalable and reproducible evaluation. For each query, a model is required to select one option (optionally accompanied by intermediate reasoning), and performance is measured using exact-match accuracy on the final answer. We split the dataset into 24,357 samples for training, 1,943 for validation, and 3,000 for testing. 

Evaluation results are reported both \emph{per task}, to analyze strengths and weaknesses across different reasoning types and disaster stages, and \emph{overall}, to measure holistic disaster-response competence across conditions. 
All evaluations follow a unified multiple-choice VQA format, with exact-match accuracy on the final answer used as the metric. The dataset is split into training, validation, and test sets to ensure reproducibility and consistency, and structured <thinking> → <answering> tags are enforced to maintain reasoning traceability across tasks.

\section{DisasterVL}
In this section, we present the model \textbf{DisasterVL} for disaster reasoning. Low-altitude UAV emergency response requires multimodal models to perform reliable multi-step reasoning from noisy and partial visual observations under tight on-site compute constraints. In practice, directly fine-tuning a small vision--language model on disaster QA is often brittle: models lack sufficient domain grounding, suffer unstable vision--language alignment, and exhibit inconsistent decision behavior on complex reasoning tasks.
Motivated by these challenges, we propose to optimize DisasterVL with a three-stage progressive training framework for lightweight multimodal disaster reasoning: (i) domain knowledge injection, (ii) multimodal alignment with structured reasoning, and (iii) policy-based refinement. 
The framework incrementally builds domain-aware semantics, grounds them in low-altitude UAV imagery with explicit reasoning structure across pre-, during-, and post-disaster stages, and refines decision behavior to improve multi-step disaster reasoning under limited computational budgets.

\subsection{Stage I: Domain Knowledge Injection}
\label{subsec:stage1}

General-purpose vision--language models lack specialized knowledge and technical terminology required for disaster analysis. We therefore construct a domain-specific text corpus by integrating geological investigation reports, historical disaster records, academic literature, and national technical standards. After cleaning and deduplication, we have 15,294 textual instruction-response pairs.
This stage aims to inject broad domain knowledge covering both foundational concepts and practical disaster-response principles, providing a semantic foundation for subsequent multimodal learning. 

Formally, let
$\mathcal{D}_{\text{stage}_1} = \{(x_i, y_i)\}_{i=1}^{N}$
denote the dataset, where $x_i$ is a domain instruction and $y_i$ is the corresponding expert response. Training minimizes the standard negative log-likelihood:
\begin{equation}
\mathcal{L}_{\text{stage}_1}(\theta)
= - \mathbb{E}_{(x,y)\sim\mathcal{D}_{\text{stage}_1}}
\left[\log p_\theta(y\mid x)\right].
\end{equation}
Optimizing this objective enables the model to internalize domain concepts, terminology, and reasoning patterns in a unified semantic space.

\subsection{Stage II: Multimodal Alignment}
\label{subsec:stage2}

With domain-aware semantics established in Stage~I, Stage~II aligns low-altitude UAV visual evidence with this textual space. We construct an image-text instruction set $\mathcal{D}_{\text{stage}_{2}}$ from the \emph{training split} of DisasterBench, where each sample pairs a UAV image with a task-conditioned instruction and an expert response.
During training, we freeze the visual encoder $f_{\mathrm{vis}}(\cdot)$ and perform visual instruction tuning. Given an input image $I$, we extract visual features as $v = f_{\mathrm{vis}}(I)$, and optimize
\begin{equation}
\mathcal{L}_{\text{stage}_{2}}(\theta)
= - \mathbb{E}_{(I,x,y)\sim\mathcal{D}_{\text{stage}_{2}}}
\left[\log p_\theta(y\mid x, v)\right],
\end{equation}
where $x$ is a task-related instruction and $y$ is the corresponding expert response, and only the language model is updated.

To encourage explicit and consistent reasoning, we adopt a chain-of-thought format that separates intermediate reasoning from the final prediction:
\begin{equation}
y = [y^{\text{think}},\, y^{\text{ans}}],
\end{equation}
\begin{equation}
p_\theta(y\mid x,v)
= p_\theta(y^{\text{think}}\mid x,v)\,
p_\theta(y^{\text{ans}}\mid x,v,y^{\text{think}}).
\end{equation}
Here, $y^{\text{think}}$ is enclosed in \texttt{<thinking>} and $y^{\text{ans}}$ is enclosed in \texttt{<answering>}. This structured interface improves format stability and provides a clean target for the refinement stage.
% \textcolor{red}{\textbf{This stage improves multi-step inference for tasks requiring contextual observation.}}

\subsection{Stage III: Policy-Based Refinement}
\label{subsec:stage3}

Despite structured multimodal alignment, residual reasoning errors may persist in complex disaster scenarios. As a lightweight refinement step, we apply Group Relative Policy Optimization (GRPO)~\cite{shao2024deepseekmath} to improve reasoning consistency without introducing an auxiliary value network.
For each query, the model samples a small group of candidate outputs and updates the policy based on relative performance within the group, with KL regularization to a fixed reference policy. Training data follows a multiple-choice format, and each output consists of a \texttt{<thinking>} segment followed by a single-option \texttt{<answering>} segment. The reward combines a strict format constraint with answer correctness:
\begin{equation}
r(o,q) = r_{\text{format}}(o) + \lambda\, r_{\text{acc}}(o,q).
\end{equation}
This stage serves as a lightweight refinement step that improves reasoning coherence while maintaining training stability and computational efficiency.
It reduces error propagation from earlier stages and enhances multi-step decision consistency.

\section{Experiments}

\subsection{Experiment Setup}

\paragraph{Baseline Models}
We evaluate \textbf{eight closed-source} MLLMs: GPT-4o, GPT-4o mini~\cite{hurst2024gpt}, GPT-5, Gemini-2.5-Pro, Gemini-2.0-Flash, Gemini-2.5-Flash~\cite{comanici2025gemini}, Claude-Sonnet-4.5, and Grok-4, where GPT-4o mini and the Gemini Flash models serve as efficiency-oriented counterparts to their flagship versions. We further include \textbf{thirteen open-source} MLLMs (1B--8B), with an emphasis on lightweight models: Qwen2-VL-2B-Instruct~\cite{wang2024qwen2}, Qwen2.5-VL (3B, 7B)~\cite{bai2025qwen2}, Janus-Pro-1B~\cite{chen2025janus}, InternVL2 (2B, 4B, 8B)~\cite{chen2024expanding}, InternVL-3.5 (1B, 2B, 8B)~\cite{wang2025internvl3}, DeepSeek-VL2 (Tiny, Small)~\cite{wu2024deepseek}, and Phi-Vision (Phi-3-Vision 4.15B)~\cite{Abdin2024Phi3TR}. More model configuration details are provided in the supplementary material.
\paragraph{Implementation and Evaluation}
We adopt \textbf{Qwen2-VL-2B-Instruct} as the base model for DisasterVL. Across all three training stages, the vision encoder remains frozen while only the language model is updated. In Stage I, the base model is fine-tuned with Low-Rank Adaptation (LoRA) on the domain-specific text corpus (rank 32, $\alpha$=64) for three epochs using Adam (learning rate $1\times10^{-4}$) with a cosine scheduler, on 4 NVIDIA L40S GPUs (global batch size 32). Stage II continues LoRA fine-tuning on the image–text instruction set with identical settings except a LoRA dropout of 0.05. Stage III applies Group Relative Policy Optimization (GRPO) for 40 steps, using AdamW (learning rate $1\times10^{-6}$, weight decay $1\times10^{-2}$) with a rollout batch size of 1024 and global batch size 256, sampling four candidate responses per query and applying KL regularization and accuracy reward ($\lambda$=1.5). For evaluation, a model may output either a single option letter or reasoning followed by a final option. We use strict exact-match parsing: a prediction is counted as correct only if the extracted final output is a \textit{single} valid option letter in the required format.

\begin{table*}[!ht]
\centering
\begin{adjustbox}{max width=\textwidth}
\begin{tabular}{lcccccccccccc}
\toprule
& \multicolumn{10}{c}{\textbf{Testing Set}} & \multicolumn{2}{c}{\textbf{Validation Set}} \\
\cmidrule(lr){2-11} \cmidrule(lr){12-13}
\textbf{Model} & \textbf{RAE} & \textbf{TMI} & \textbf{SUC} & \textbf{PPI} & \textbf{CBA} & \textbf{LSA} & \textbf{CRR} & \textbf{ERA} & \textbf{DCS} & \textbf{ Overall} & \textbf{Overall} & \textbf{\#Token} \\
\cmidrule(lr){1-11} \cmidrule(lr){12-13}
\multicolumn{12}{c}{\textit{Closed-Source Models}} \\
\cmidrule(lr){1-11} \cmidrule(lr){12-13}
GPT-4o mini & 27.19\% & 64.86\% & 81.76\% & 63.39\% & 55.03\% & 55.27\% & 62.44\% & 83.19\% & 73.19\% & 64.00\% & 67.27\% & 423 \\ 
GPT-4o & 40.35\% & 76.81\% & 86.32\% & 58.33\% & 53.85\% & 76.37\% & 71.95\% & 83.19\% & 79.85\% & 73.00\% & 73.19\% & 335\\
GPT-5 & 66.67\% & 84.06\% & 92.51\% & 78.57\% & 74.56\% & 87.06\% & 80.09\% & 97.48\% & 90.11\% & 84.83\% & 83.69\% & 644\\
Gemini-2.5-pro & 58.77\% & 84.06\% & 94.14\% & 76.49\% & 70.41\% & 87.06\% & 80.32\% & 93.28\% & 90.49\% & 84.17\% & 87.70\% & 1323 \\
Gemini-2.0-flash & 47.37\% & 76.45\% & 89.90\% & 75.89\% & 65.68\% & 71.17\% & 81.45\% & 88.24\% & 92.40\% & 78.80\% & 81.68\% & 435 \\
Gemini-2.5-flash & 55.26\% & 83.33\% & 92.83\% & 63.10\% & 69.82\% & 77.64\% & 71.04\% & 91.60\% & 87.07\% & 78.03\% & 82.96\% & 1321 \\
Claude-Sonnet-4-5 & 52.63\% & 80.07\% & 91.86\% & 68.45\% & 71.01\% & 82.28\% & 76.02\% & 93.28\% & 86.12\% & 79.93\% & 86.26\% & 457 \\
Grok-4 & 53.51\% & 76.09\% & 89.58\% & 76.19\% & 62.72\% & 81.15\% & 69.68\% & 89.92\% & 82.51\% & 77.80\% & 77.87\% & 733 \\
\cmidrule(lr){1-11} \cmidrule(lr){12-13}
\multicolumn{12}{c}{\textit{Open-Source Models}} \\
\cmidrule(lr){1-11} \cmidrule(lr){12-13}
Qwen2-VL-2B & 13.16\% & 42.75\% & 69.06\% & 52.98\% & 24.26\% & 22.78\% & 44.57\% & 69.75\% & 58.94\% & 43.87\% & 49.92\% & 352 \\
Qwen2.5-VL-3B & 47.37\% & 55.43\% & 75.57\% & 62.20\% & 34.32\% & 42.33\% & 63.80\% & 89.08\% & 75.67\% & 59.77\% & 65.31\% & 428 \\
Qwen2.5-VL-7B & 56.14\% & 67.03\% & 85.02\% & 72.32\% & 40.83\% & 63.29\% & 70.14\% & 92.44\% & 86.69\% & 71.60\% & 74.94\% & 425 \\
Janus-Pro-1B & 14.91\% & 43.84\% & 76.55\% & 53.57\% & 28.99\% & 57.52\% & 34.62\% & 60.50\% & 61.41\% & 51.97\% & 55.12\% & 354 \\
InternVL2-2B & 9.65\% & 52.17\% & 77.52\% & 46.43\% & 29.59\% & 33.33\% & 50.68\% & 76.47\% & 63.12\% & 49.43\% & 54.09\% & 425 \\
%InternVL2-4B & 28.95\% & 57.97\% & 85.02\% & 70.54\% & 35.50\% & 51.90\% & 53.17\% & 88.24\% & 80.61\% & 62.80\% & 68.71\% & 146 \\
InternVL2-8B & 61.40\% & 63.77\% & 81.43\% & 65.18\% & 42.60\% & 53.87\% & 72.40\% & 92.44\% & 84.98\% & 68.23\% & 71.38\% & 255 \\
InternVL-3-5-1B & 12.28\% & 33.70\% & 80.13\% & 38.99\% & 40.83\% & 49.65\% & 31.22\% & 48.74\% & 50.57\% & 45.60\% & 55.79\% & 380 \\
InternVL-3-5-2B & 30.70\% & 45.29\% & 82.08\% & 50.89\% & 41.42\% & 60.76\% & 46.83\% & 67.23\% & 73.76\% & 58.67\% & 66.13\% & 355 \\
InternVL-3-5-4B & 51.75\% & 55.80\% & 82.08\% & 57.14\% & 59.76\% & 72.86\% & 54.30\% & 85.71\% & 78.14\% & 67.63\% & 72.41\% & 459 \\
InternVL-3-5-8B & 43.86\% & 54.35\% & 83.71\% & 61.31\% & 55.62\% & 68.64\% & 51.36\% & 88.24\% & 78.90\% & 66.40\% & 72.31\% & 322 \\
Deepseek-VL2-T & 13.16\% & 38.41\% & 78.50\% & 40.18\% & 36.69\% & 54.85\% & 39.37\% & 46.22\% & 39.73\% & 46.23\% & 52.91\% & 190 \\
Deepseek-VL2-S & 31.58\% & 54.71\% & 92.51\% & 61.31\% & 53.25\% & 64.56\% & 61.54\% & 88.24\% & 79.47\% & 67.37\% & 70.97\% & 359 \\
Phi-3-Vision(4.15B) & 31.58\% & 59.42\% & 83.71\% & 74.70\% & 31.95\% & 44.59\% & 67.19\% & 84.03\% & 83.84\% & 63.90\% & 66.80\% & 272 \\
%Phi-4-Vision(6B) & 43.86\% & 55.80\% & 83.71\% & 61.01\% & 39.05\% & 58.93\% & 64.48\% & 88.24\% & 82.32\% & 65.80\% & 71.69\% & 143\\
\cmidrule(lr){1-11} \cmidrule(lr){12-13}
\textbf{DisasterVL (Ours-2B)} & 50.88\% & 57.97\% & 91.86\% & 80.95\% & 51.48\% & 70.04\% & 59.05\% & 85.71\% & 87.07\% & 72.60\% & 81.57\% & 168 \\
\bottomrule
\end{tabular}
\end{adjustbox}
\caption{Performance of closed- and open-source vision-language models on the \textit{test} set and \textit{val} set of DisasterBench. We report task-wise accuracy and overall accuracy. $\#Tokens$ denotes the average number of generated tokens per sample, reflecting inference efficiency.}
% \vspace{-0.3cm}
\label{tab:bench_all}
\end{table*}

\subsection{Performance Comparison}
\Cref{tab:bench_all} presents a comprehensive evaluation of \textit{DisasterVL} as well as a wide range of state-of-the-art closed- and open-source vision-language models on \textit{DisasterBench}. On the test split, DisasterVL achieves 72.60\% overall accuracy, representing the strongest performance among lightweight open-source models. In particular, it substantially outperforms same-scale baselines, exceeding Qwen2-VL-2B (43.87\%) by +28.73 points and InternVL2-2B (49.43\%) by +23.17 points in overall accuracy. Despite its compact size, DisasterVL is also highly parameter-efficient, surpassing larger open-source models such as Qwen2.5-VL-7B (71.60\%) and InternVL2-8B (68.23\%), as well as the efficient closed-source GPT-4o mini (64.00\%).
Beyond aggregate accuracy, \textit{DisasterVL} exhibits balanced and robust performance across tasks, particularly on reasoning-intensive categories that require multi-step inference and causal understanding. Notably, it achieves strong results on Propagation Path Inference (80.95\%), Situation Understanding and Classification (91.86\%), and Disaster-Chain Summary (87.07\%), demonstrating its effectiveness in modeling hazard evolution, contextual interpretation, and holistic disaster reasoning from low-altitude UAV imagery.
These trends are consistent on the validation split, where DisasterVL attains 81.57\% overall accuracy, again outperforming comparable-scale backbones while remaining competitive with substantially larger models. 

In addition to accuracy, DisasterVL is token-efficient, generating only 168 tokens on average, which is significantly fewer than most closed-source models (e.g., GPT-4o mini 423; Gemini-2.5-Pro 1323) and strong open-source models (e.g., Qwen2.5-VL-7B 425). This efficiency directly supports practical deployment under on-site and edge-compute constraints common in real-world emergency response scenarios. The close alignment between validation and test performance further indicates that the observed gains are not attributable to overfitting, but rather arise from improved multimodal reasoning and decision consistency enabled by our three-stage training framework. Qualitative results can be found in the supplementary materials.

To further investigate the impact of Stage III’s policy-based refinement, we evaluated nearby values of the accuracy-weighting coefficient $\lambda$. The overall test accuracy observed for different $\lambda$ values was 69.93\% ($\lambda$=2), 72.60\% ($\lambda$=1.5), 70.47\% ($\lambda$=1), and 70.73\% ($\lambda$=0.5), indicating that $\lambda$=1.5 provides the best trade-off between reward scaling and model performance.
The model’s results at each stage are presented in the supplementary materials.

\begin{table*}[t]
\centering
\begin{adjustbox}{max width=\textwidth}
\begin{tabular}{lccccccccccccccc}
\toprule
Stage & Lahar & Explosion & Rock fall & Ice-jam flood & Fissures & Debris flow & Snow avalanche & Lava flow & Pyroclastic flow & Subsidence & Landslide & Hurricane & Fire & Flood & Overall \\
\midrule
Stage1       & 73.97\% & 80.57\% & 70.75\% & 60.52\% & 51.95\% & 59.82\% & 50.00\% & 54.20\% & 69.23\% & 65.15\% & 64.64\% & 50.63\% & 49.00\% & 70.00\% & 59.67\% \\
+Stage2      & 80.82\% & 84.00\% & 77.55\% & 62.66\% & 51.95\% & 69.64\% & 59.18\% & 63.36\% & 70.38\% & 62.12\% & 65.36\% & 55.70\% & 60.45\% & 62.14\% & 64.13\% \\
+Stage3 (Ours) & 87.67\% & 89.14\% & 81.63\% & 70.82\% & 57.14\% & 80.36\% & 69.39\% & 70.99\% & 80.38\% & 75.76\% & 71.79\% & 62.34\% & 67.66\% & 76.79\% & 72.60\% \\
% \midrule
\bottomrule
\end{tabular}
\end{adjustbox}
\caption{Accuracy of different training stages across disaster categories}
% \vspace{-0.3cm}
\label{tab:ab_category_wise}
\end{table*}

\subsection{Method Analysis}
\paragraph{Stage-wise Performance}
\Cref{tab:ab_category_wise} presents the test-set accuracy of DisasterVL across different disaster categories for each training stage. Stage~1 (domain knowledge injection) provides a strong baseline, particularly on knowledge-heavy categories, while Stage~2 (multimodal alignment with structured reasoning) consistently improves perception-driven categories like Debris Flow and Snow Avalanche. The full three-stage pipeline (+Stage3) achieves the highest accuracy across nearly all disaster types, demonstrating complementary effects: Stage~3 (policy-based refinement) particularly enhances multi-step reasoning tasks. Overall, the progressive curriculum stabilizes performance across rare and frequent categories alike, with an overall test accuracy of 72.6\%, confirming that the staged training effectively combines domain grounding, visual reasoning, and decision refinement.

\begin{table}[t]
\centering
\footnotesize
\setlength{\tabcolsep}{10pt} % 控制列间距

\begin{tabular*}{\linewidth}{@{\extracolsep{\fill}}ccccc}
\toprule
\textbf{Stage I} & \textbf{Stage II} & \textbf{Stage III} & \textbf{Val Acc.} & \textbf{Test Acc.} \\
\midrule
            &               &               &43.87\%  &49.92\%  \\
\midrule
\checkmark  &               &               &69.38\%  &59.67\%  \\
            &\checkmark     &               &72.26\%  &66.43\% \\
            &               &\checkmark     &69.48\%  &61.53\%  \\
\checkmark  &\checkmark     &               &72.98\%  &64.13\%  \\
\checkmark  &               &\checkmark     &76.38\%  &67.07\%  \\
            &\checkmark     &\checkmark     &75.81\%  &68.03\%  \\
\midrule
\checkmark & \checkmark     &\checkmark     &81.57\%  &72.60\%  \\
\bottomrule
\end{tabular*}

\caption{Impact of different training stages.}
\label{tab:ablation_component}
\end{table}

\paragraph{Component Ablation}
\Cref{tab:ablation_component} shows that each stage contributes meaningfully and that the three stages are complementary. The base model without specialized training performs poorly (49.92\% test). Applying a single stage improves accuracy but remains insufficient: Stage~I reaches 59.67\% test, Stage~II 66.43\% test, and Stage~III 61.53\% test. Combining two stages yields further gains (64.13\%--68.03\% test), indicating that semantic grounding, visual alignment, and decision refinement address different failure modes. The full pipeline achieves the best results, reaching 81.57\% validation and 72.60\% test accuracy, confirming that robust disaster reasoning in lightweight models benefits from all three stages.

\begin{table}[t]
\centering
\footnotesize
\setlength{\tabcolsep}{5pt}

\begin{tabular*}{\linewidth}{@{\extracolsep{\fill}}lcccc}
\toprule
Task & Stage1 & +Stage2 & +Stage3 (Ours) & Full-SFT \\
\midrule
RAE     & 34.21\%  & 35.96\%   & 50.88\% & 48.25\% \\
TMI     & 56.88\%  & 53.26\%   & 57.97\% & 52.90\% \\
SUC     & 88.60\%  & 87.62\%   & 91.86\% & 88.60\% \\
PPI     & 59.52\%  & 66.67\%   & 80.95\% & 76.79\% \\
CBA     & 38.18\%  & 40.83\%   & 51.48\% & 49.70\% \\
LSA     & 43.60\%  & 60.06\%   & 70.04\% & 67.93\% \\
CRR     & 53.85\%  & 49.10\%   & 59.05\% & 52.26\% \\
ERA     & 83.19\%  & 82.35\%   & 85.71\% & 84.87\% \\
DCS     & 80.61\%  & 82.13\%   & 87.07\% & 86.12\% \\
\midrule
Overall & 59.67\%  & 64.13\%   & 72.60\% & 69.43\% \\
\bottomrule
\end{tabular*}

\caption{Comparison between three stages and Full-SFT}
\label{tab:ablation_full_sft}
\end{table}

\paragraph{Stage and Full Training Comparison}
\Cref{tab:ablation_full_sft} compares DisasterVL’s performance across the three training stages with a single-stage Full-SFT baseline. Stage~I establishes a reasonable foundation, while Stage~II improves perception-dependent tasks. Incorporating Stage~III consistently boosts accuracy across nearly all tasks, notably on multi-step reasoning tasks. Compared with Full-SFT, the full three-stage pipeline achieves higher overall accuracy (72.60\% vs. 69.43\%), demonstrating that the progressive curriculum provides more stable and effective disaster reasoning than direct end-to-end SFT.

\paragraph{Effect of Training Order}
\Cref{tab:ablation_component_order} reports the impact of different training orders for the three stages. While most permutations yield moderate improvements over single-stage training (65.10\%--69.13\% test accuracy), none match the performance of the proposed pipeline. In particular, applying Stage~II or Stage~III before domain knowledge injection (Stage~I) consistently leads to suboptimal results, suggesting that multimodal alignment and policy refinement require a strong semantic foundation. The canonical order \emph{Stage~I $\rightarrow$ Stage~II $\rightarrow$ Stage~III} achieves the highest performance (81.57\% validation, 72.60\% test), confirming that progressively injecting domain knowledge, grounding it in visual reasoning, and finally refining decision policies is critical for effective lightweight disaster reasoning.

\begin{table}[t]
\centering
\footnotesize
\setlength{\tabcolsep}{5pt}

\begin{tabular*}{\linewidth}{@{\extracolsep{\fill}}ccc}
\toprule
\textbf{Training Order} & \textbf{Val Acc.} & \textbf{Test Acc.} \\
\midrule
Stage I $\rightarrow$ Stage III $\rightarrow$ Stage II     &79.52\%  &68.73\%  \\
Stage II $\rightarrow$ Stage I $\rightarrow$ Stage III     &66.23\%  &65.10\%  \\
Stage II $\rightarrow$ Stage III $\rightarrow$ Stage I     &76.84\%  &68.03\%  \\
Stage III $\rightarrow$ Stage I $\rightarrow$ Stage II     &79.72\%  &67.70\%  \\
Stage III $\rightarrow$ Stage II $\rightarrow$ Stage I     &78.13\%  &69.13\%  \\
\midrule
Stage I $\rightarrow$ Stage II $\rightarrow$ Stage III     &81.57\%  &72.60\%  \\
\bottomrule
\end{tabular*}

\caption{Impact of the order of the three stages.}
\label{tab:ablation_component_order}
\end{table}

\paragraph{Backbone Generalization}
% original version
\Cref{tab:backbone} evaluates whether our training pipeline generalizes across different lightweight backbones. Applying our method consistently yields large gains on both validation and test splits: for example, Qwen2-VL-2B improves from 43.87\% to 72.60\% on the test set, and InternVL-3.5-1B improves from 45.60\% to 68.03\%. Similar improvements are observed for stronger backbones such as Qwen2.5-VL-3B (59.77\% $\rightarrow$ 72.80\%) and InternVL-3.5-2B (58.67\% $\rightarrow$ 72.07\%), and notably also scale to larger models like Qwen2.5-VL-7B (71.60\% $\rightarrow$ 79.80\%). Beyond accuracy, our models are also more token-efficient, reducing average generation from 352 to 168 tokens on Qwen2-VL-2B, with consistent reductions across other backbones. These results indicate that our three-stage framework provides robust, transferable improvements rather than relying on a specific architecture.

\begin{table}[t]
\centering
\footnotesize
\setlength{\tabcolsep}{5pt} % 控制列间距

\begin{tabular*}{\linewidth}{@{\extracolsep{\fill}}lccc}
\toprule
\textbf{Backbone} & \textbf{Val Acc.} & \textbf{Test Acc.} & \textbf{\#Token} \\
\midrule
Qwen2-VL-2B     & 49.92\% & 43.87\% & 352 \\
+ Ours          & 81.57\% & 72.60\% & 168 \\
\midrule
Qwen2.5-VL-3B   & 65.31\% & 59.77\% & 428 \\
+ Ours          & 79.67\% & 72.80\% & 202 \\
\midrule
Qwen2.5-VL-7B   & 74.94\% & 71.60\% & 425 \\
+ Ours          & 86.67\% & 79.80\% & 232 \\
\midrule
InternVL-3.5-1B & 55.79\% & 45.60\% & 380 \\
+ Ours          & 76.12\% & 68.03\% & 205 \\
\midrule
InternVL-3.5-2B & 66.13\% & 58.67\% & 355 \\
+ Ours          & 80.55\% & 72.07\% & 224 \\
\bottomrule
\end{tabular*}

\caption{Validation overall accuracy with different backbones. The upper block reports the base backbones, and the lower block reports the corresponding models trained with our pipeline.}
\label{tab:backbone}
\end{table}

\begin{figure}[t]
    \centering
    \includegraphics[width=1\linewidth]{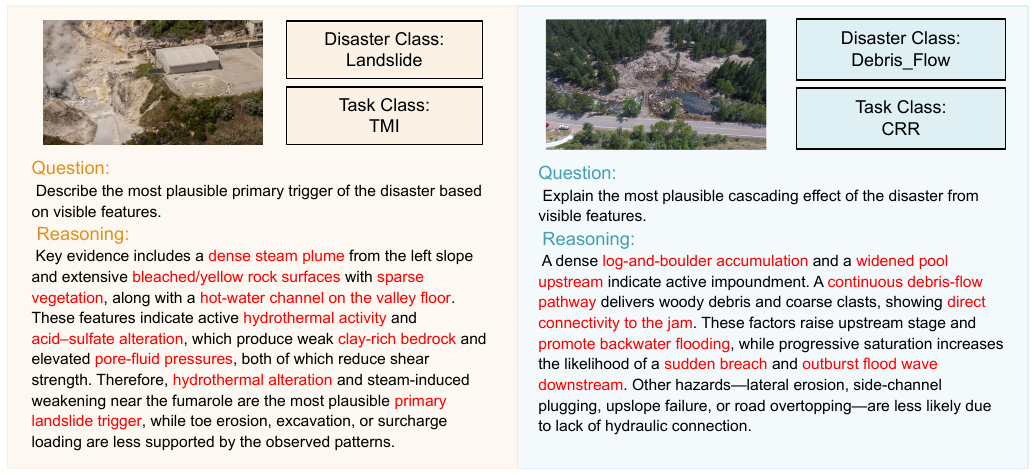}
    \caption{Examples of open-ended reasoning on DisasterBench images, the red-highlighted keywords are core visual cues extracted from UAV imagery, serving as the logical foundation for multimodal reasoning.}
    \label{fig:fulldataset_statistics}
    % \vspace{-0.4cm}
\end{figure}

\paragraph{Importance of Visual Information}
To rigorously evaluate DisasterBench’s dependence on visual reasoning, we perform a text-only ablation in which models are provided solely with questions and answer choices. The removal of visual inputs leads to a substantial drop in performance (e.g., from 72.6\% to 57.9\% overall), demonstrating that visual evidence is critical for accurate disaster-related reasoning. Representative reasoning instances, as illustrated in Fig.~\ref{fig:fulldataset_statistics}, highlight how models integrate spatial patterns, structural cues, and feature distributions from low-altitude UAV images with task-conditioned questions to generate coherent reasoning and final answers. For instance, models analyze slope features and surface patterns to infer potential causes of a landslide, and they track the connectivity and accumulation of debris to anticipate cascading effects downstream. These examples emphasize that multi-step disaster inference relies heavily on interpreting visual information and systematically connecting observations to outcomes, underscoring the importance of maintaining rich visual context for effective decision-making in the multi-stage reasoning tasks.

\section{Conclusion}

We introduced DisasterBench, a multi-stage multimodal reasoning benchmark for UAV-Based disaster response in complex environments. DisasterBench covers diverse disaster types and response-critical tasks across pre-, during-, and post-disaster stages, explicitly evaluating reasoning beyond perception, including causal analysis, hazard evolution, damage understanding, and decision-oriented assessment. Built from real-world UAV imagery with expert verification, it provides a realistic testbed for studying stage-aware multimodal reasoning in emergency settings. To support deployment under on-site compute constraints, we further proposed DisasterVL, a lightweight multimodal model trained with a three-stage framework that combines domain knowledge injection, chain-of-thought-guided multimodal alignment, and policy-based refinement. Experiments across 21 closed- and open-source MLLMs show that DisasterVL achieves strong and balanced performance among lightweight models while remaining token-efficient.

\clearpage

\clearpage

\bibliographystyle{elsarticle-num} 
\bibliography{custom}

\end{document}